\documentclass[conference]{IEEEtran}
\IEEEoverridecommandlockouts

\usepackage[ backend=biber, style=ieee,
            maxbibnames=1, minbibnames=1]{biblatex}
\usepackage{bm}
\usepackage{amsmath,amssymb,amsfonts}
\usepackage[ruled, vlined, linesnumbered]{algorithm2e}
\usepackage{graphicx}
\usepackage{float}
\usepackage{caption}
\usepackage{subcaption}
\usepackage{textcomp}
\usepackage{xcolor}
\usepackage{marvosym}
\usepackage{multirow}
\usepackage{booktabs}
\usepackage{array}
\usepackage{url}
\usepackage{enumitem}
\usepackage{xcolor}
\usepackage[hidelinks]{hyperref}
\def\BibTeX{{\rm B\kern-.05em{\sc i\kern-.025em b}\kern-.08em
		T\kern-.1667em\lower.7ex\hbox{E}\kern-.125emX}}
\IEEEoverridecommandlockouts

\begin{document}
\title{Cooperative Task Offloading through Asynchronous Deep Reinforcement Learning in Mobile Edge Computing for Future Networks\\	
 
	}
	
	\author{\IEEEauthorblockA {Yuelin Liu, Haiyuan Li, Xenofon Vasilakos, Rasheed Hussain, and Dimitra Simeonidou\\ }
    \IEEEauthorblockA{High Performance Networks (HPN) Research Group, Smart Internet Lab, University of Bristol, Bristol, UK\\
    Email: \{name\}.\{surname\}@bristol.ac.uk}
	
    \thanks{This work was supported by Project REASON, a UK Government funded project sponsored by the Department of Science Innovation and Technology (DSIT). This work was also supported by the Future Telecoms Research Hub, Platform for Driving Ultimate Connectivity (TITAN), 
sponsored by DSIT and the Engineering and Physical Sciences Research Council (EPSRC) under Grant EP/X04047X/2 and Grant EP/Y037243/1.}}

	\maketitle
	
\begin{abstract}

Future networks (including 6G) are poised to accelerate the realisation of Internet of Everything. The latter will imply a high demand for computational resources to support new services. Mobile Edge Computing (MEC) is a promising solution that enables offloading \textit{computation-intensive} tasks to nearby edge servers from the end-user devices, thereby \textit{reducing latency} and \textit{energy consumption}. Nevertheless, relying solely on a single MEC server for task offloading can lead to \textit{uneven resource utilisation} and suboptimal performance in complex scenarios. Additionally, traditional task offloading strategies specialise in centralised policy decisions, which unavoidably entails extreme transmission latency and reach computational bottleneck. To address these gaps, we propose a \textit{latency-efficient and energy-efficient Cooperative Task Offloading} framework with \textit{Transformer-driven Prediction} (CTO-TP), leveraging asynchronous multi-agent deep reinforcement learning to address these challenges. This approach fosters edge-edge cooperation and decreases the synchronous waiting time by performing asynchronous training, optimising task offloading, and resource allocation across distributed networks. The performance evaluation demonstrates that the proposed CTO-TP algorithm reduces up to 80\%  overall system latency and 87\% energy consumption compared to the baseline schemes.

\end{abstract}
	
	
\section{Introduction and Related Work}

The advancement of 5G communication technologies, along with current efforts towards 6G and beyond, has led to the possibility of realising exciting new services. These include, but are not limited to, real-time gaming, Augmented Reality (AR), Vehicle-to-Everything (V2X) communication, robotic surgery, and the metaverse. These services require enhanced Quality of Experience (QoE) for users and huge computing capacity. At the same time, requirements for such services easily surpass the capabilities of end-user devices in terms of computing capacity and energy. 
The most common approach to address this limitation is acquiring resources from cloud servers; however, accessing cloud computing will impose extra delays. To this end, Mobile Edge Computing (MEC) is recognised as a promising solution to address this challenge \cite{yang2021coalitional}. MEC facilitates end-user devices, including mobile devices, to process computation-intensive tasks on nearby edge servers, which can significantly reduce the devices' service latency and energy consumption \cite{gao2023large}. These devices could be even Internet of Things (IoT) devices that are constrained by energy and computational power. Benefiting from MEC, the user QoE can be dramatically improved by offloading high computational tasks to edge servers with more computational resources compared to end-user devices~\cite{liu2021efficient}. Nevertheless, the geographical distribution of MEC servers in reality is a challenge. Furthermore, the highly dynamic changes in user activities using services through end-user devices cause an uneven distribution of workload requests, both in terms of time and location. This imbalance results in varying load levels among MEC servers, leading to decreased overall resource utilisation. Some MEC servers might become fully loaded while others may remain idle. Notably, a single MEC server is often resource-limited and unable to handle peaks in task requests. Therefore, offloading tasks to distributed MEC servers without collaboration can lead to uneven resource utilisation, making it difficult to efficiently meet the large, complex, and dynamic demands of computing tasks.

To address these challenges, cooperative task offloading has recently emerged as a promising solution where distributed MEC servers can collaborate more effectively. Given the mobility of users 
and the variability over time, computational tasks can be offloaded from an overloaded edge server to a nearby server with available resources, which may be idle or less busy by comparison \cite{ren2022efficient}. However, the dynamic and complex environment, including fluctuating task requests, network resources, and multi-objective optimisation factors such as latency and energy consumption, presents challenges for achieving effective long-term optimisation.
Fortunately, AI-based strategies, especially Deep Reinforcement Learning (DRL) \cite{he2020qoe,dai2021asynchronous}, present effective solutions to address these challenges. The DRL-based cooperative task offloading strategies can effectively reduce the overload conditions and excessive resource utilisation on individual MEC servers and improve long-term computational efficiency. To this end, synchronous DRL-based methods \cite{he2020qoe} focus on having all DRL agents collaborate to train and update simultaneously, thereby enhancing training stability. However, due to varying spatiotemporal computational task demands \cite{tranos2015mobile}, task arrivals and requirements for each agent are often imbalanced. For example, some agents may remain constantly busy, while others experience long periods with no task arrivals. This imbalance causes agents to waste considerable time waiting for others to synchronise. On the other hand, asynchronous DRL-based algorithms \cite{dai2021asynchronous} reduce communication overhead and training time by allowing each agent to train independently. However, this approach prevents agents from sharing essential information, such as states and actions, during asynchronous training, which overlooks the importance of strong collaboration in cooperative task offloading.



To fill the identified gaps, in this paper, we propose a joint latency and energy-efficient Cooperative Task Offloading with Transformer-driven Prediction (CTO-TP) framework, based on asynchronous multi-agent DRL. The main contributions of this paper can be summarized as follows: 

    
%

\begin{itemize}[leftmargin=*]
    \item We model \textit{cooperative task offloading} and \textit{resource allocation} in \textit{distributed scenarios} as a Markov Decision Process (MDP) to optimise long-term latency and energy consumption, considering the impact of past actions.
    
    \item  We address the problem of \textit{dynamic task arrivals} and \textit{requirements} by adapting offline-trained transformer models tailored to predict future task arrival times and resource demands for each MEC server.
    
    \item  We present and evaluate against benchmark models a \textit{novel hybrid solution backed by Multi-Agent Deep Q Network (MADQN)} and Multi-Agent Deep Deterministic Policy Gradient (MADDPG) models, which enables collaborative interaction and asynchronous training, reducing waiting time, improving efficiency, and enhancing resource utilisation. 
    Our approach is supported by an asynchronous multi-agent DRL approach, combining MADQN for discrete server selection and MADDPG for continuous offloading and resource allocation.
\end{itemize}

The remainder of the paper is organised as follows. Section \ref{sec:sysmodel} describes the system model and problem formulation. Section \ref{sec:proposed} details the latency and energy-efficient CTO-TP framework based on the asynchronous hybrid MADQN-MADDPG. The performance evaluation is discussed in Section \ref{sec:results} followed by conclusions and future work in Section \ref{sec:conclusion}.

\section{System Model}
\label{sec:sysmodel}

\begin{figure}[t]  
    \centering
    \includegraphics[width=0.4\textwidth]{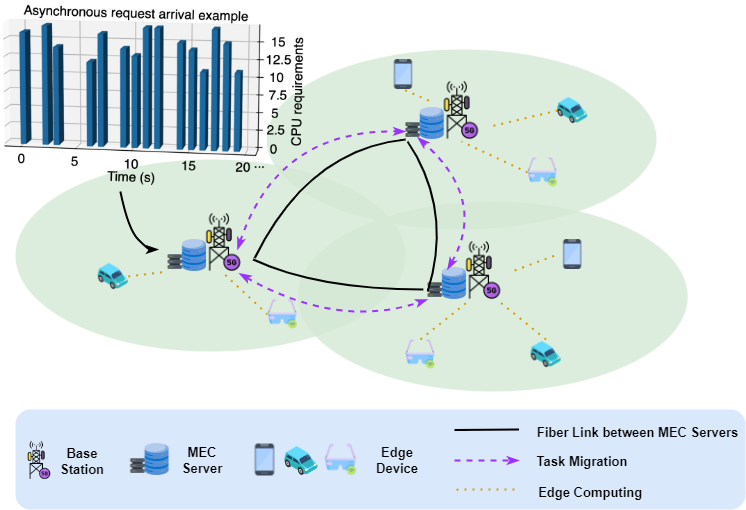}
    \caption{Cooperative MEC network with asynchronous task request arrival.}
    \label{fig:MECnetwork}

\end{figure}

In this section, we discuss our system model and problem formulation. As shown in Fig.\ref{fig:MECnetwork}, a cooperative MEC network comprising $M$ MEC servers with asynchronous task request arrival is considered, where each base station is equipped with a MEC server and is connected to other base stations by fibre links so that tasks can be migrated to other MEC servers for collaborative processing. The task arrivals at each MEC server are asynchronous, i.e., the task arrival time is independent across different MEC servers. This asynchrony reflects the real-world scenario where task requests are influenced by service users' activities, leading to imbalanced and heterogeneous task distribution among MEC servers.
\vspace{-0.2cm}
 \subsection{Communication Model}
In the considered cooperative MEC environment, MEC servers communicate with each other to share computational resources effectively and process tasks collaboratively. This will result in optimising computational resource utilisation to minimise overall system latency. Each MEC server, associated with a base station, can offload tasks to other cooperative MEC servers through existing fibre links. This inter-MEC communication will alleviate processing overhead by distributing the computational workload across multiple servers. The communication model involves data transmission between local MEC servers and cooperative MEC servers. According to \cite{schwartz2005mobile}, data rate $p_{mnt}$ on link $k$ is simplified to:
\vspace{-0.1cm}
        \begin{equation}
            p_{mnt} = b_{mnt} \log_{2} (1 + \eta)
        \end{equation}
        where $\eta$ and $b_{mnt}$ are the signal-to-noise ratio and allocated bandwidth on link $k$ from local MEC server $m$ to cooperative MEC server $n$ at time slot $t$, respectively. Based on the data rate $p_{mnt}$ and workload size $F_{mnt}$ on link $k$, the latency of offloading transmission from local MEC node \( m \) to cooperative MEC node \( n \) can be achieved by:
\vspace{-0.1cm}
        \begin{equation}
        L_{mnt}^{\text{trans}} = F_{mnt} / p_{mnt}
        \end{equation}

\subsection{Computation Model}
In the cooperative MEC network, each MEC server is responsible for handling computational tasks that arrive asynchronously. The computing workload can be processed locally or offloaded to a cooperative MEC server for effective load balancing. 

\begin{itemize}
    \vspace{-0.1cm}
    \item \ \textit{Local MEC computing}: Computational tasks are processed directly at the local MEC server where they originate. The latency $L_{mt}^{\text{compu}}$ involved in computation in local MEC node \( m \) at time slot \( t \) depends on both the required computing resources $D_{mt}$ for the arriving tasks and the allocated resource $d_{mt}$ on local MEC $m$. $L_{mt}^{\text{compu}}$ is represented by:
\vspace{-0.1cm}
        \begin{equation}
         L_{mt}^{\text{compu}} = D_{mt} / d_{mt}
        \end{equation}
    \vspace{-0.6cm}
    \item \textit{Cooperative MEC computing}: In addition to local processing, the model also incorporates cooperative MEC computing, wherein the arriving computational tasks can be offloaded from local MEC server \( m \) to other cooperative MEC server \( n \). This process not only helps in balancing the load but also contributes to reducing the response time and energy consumption for processing tasks. The computation latency $L_{mnt}^{\text{compu}}$ for cooperative MEC computing depends on the required computing resource $D_{nt}$ for offloaded tasks from local MEC $m$ to cooperative MEC $n$ and the allocated resource $d_{nt}$ at the cooperative MEC server \( n \).
    The latency of computation in cooperative MEC node \( n \) is given by:
\vspace{-0.15cm}
        \begin{equation}
         L_{mnt}^{\text{compu}} = D_{nt} / d_{nt}
        \end{equation}
\vspace{-0.1cm}
    The energy consumption $E_{mt}$ of MEC $m$ can be expressed as:
\vspace{-0.15cm}
        \begin{equation}
        E_{mt} = P_{min} T_{mt} + (P_{max} - P_{min}) U_{mt} T_{mt},
        \end{equation}
\vspace{-0.1cm}
    where $P_{min}$ and $P_{max}$ represent the power of the MEC server at zero load and full load, respectively. $U_{mt}$ and $T_{mt}$ are the computational resource utilisation rate and maximum computing time of MEC $m$ at time slot $t$, respectively. This model is adapted from \cite{ali2020deep}, where a linear relationship between power and load is assumed.

\end{itemize}    
\subsection{Problem Formulation}
In the cooperative MEC network environment, our objective is to optimise overall system performance by minimising the total cost associated with latency and energy consumption. Each task has the potential to be processed by a combination of using local computational resources and offloading tasks to a cooperative MEC server. However, the above introduces both local computation delays and offloading latency which comprises transmission delay and the computational processing time required at a remote MEC server.
Therefore, total system latency $L_{\text{total}}$ at time slot $t$ involves computational delay and transmission latency, which is represented as:
    \vspace{-0.15cm}
    \begin{equation}
    L_{\text{total}} = \sum_{m=1}^{M} \max \left\{ L_{mt}^{\text{compu}},\ L_{mnt}^{\text{trans}} + L_{mnt}^{\text{compu}}\right\}
    \end{equation}
\vspace{-0.1cm}
    In addition to latency, energy consumption is another key factor in MEC environments, where each MEC server needs to maintain efficient resource utilisation to ensure sustainability. The total energy consumption of the system at time slot $t$ is:
\vspace{-0.1cm}
    \begin{equation}
    E_{\text{total}} = \sum_{m=1}^{M}\ E_{mt}
    \end{equation}
\vspace{-0.15cm}
    Given the above, the objective function below captures the minimisation of the combined costs of delay and energy consumption:
\vspace{-0.2cm}

\begin{subequations}
\begin{equation}
     \quad \min \sum_{t=1}^{T} \left( \lambda L_{\text{total}} + \rho E_{\text{total}} \right) \label{objective} \\
\end{equation}
\vspace{-3.4ex}
\begin{align}
    \text{s.t.} \quad & \lambda + \rho = 1, \lambda \in (0,1), \rho \in (0,1) \label{C1} \\
    & d_{mt} \leq C_{mt}, d_{nt} \leq C_{nt}, \forall m, n \in M,\ m \neq n, \forall t \in T \label{C2} \\
    & b_{mnt} \leq B_{kt}, \forall m, n \in M, m \neq n, \forall k \in K_t, \forall t \in T \label{C3} \\
    & 0 \leq a_{mdt} \leq 1, \forall m \in M, \forall t \in T \label{C4}
\end{align}
\end{subequations}

where $\lambda$ and $\rho$ are balancing factors that adjust the relative contributions of delay and energy consumption to the total cost. The parameter 
$\lambda$, measured in $\text{sec}^{-1}$, modulates the impact of delay, while $\rho$, in $\text{joule}^{-1}$, scales the contribution of energy consumption. 
These factors enable the formulation of a unified cost metric, ensuring that both delay and energy are appropriately weighted in the optimisation process. Also, the resulting combined cost is \textit{unitless}, which can be desirable for formulating the optimisation problem. 
If not, the total cost can be easily turned into monetary by expressing $\lambda$ in $\frac{\$}{\text{sec}}$ and $\rho$ in $\frac{\$}{\text{joule}}$.
 $C_{mt}$, $C_{nt}$ and $B_{kt}$ are the remaining resources in local MEC $m$, remote MEC $n$, and the link between them at time $t$, respectively. Action $a_{mdt}$ is the task offloading ratio. 
$T$ can be regarded as an indefinitely extended time scale. Constraints \eqref{C2} and \eqref{C3} guarantee that both allocated computational and bandwidth resources \textit{never} exceed the remaining resources, while  constraint \eqref{C4} ensures that the action of task offloading ratio is maintained within the range of 0 to 1.

\section{Proposed CTO-TP Framework}
\label{sec:proposed}
To jointly optimise the cooperative task offloading and resource allocation decisions, we refine the problem as a multi-agent MDP and address it by applying an asynchronous transformer-driven hybrid algorithmic combination of MADQN and MADDPG across distributed networks. The MADQN algorithm, which operates in a discrete action domain, is used to facilitate effective offloading decisions, while MADDPG provides precise control of resource allocation within a continuous action space. Furthermore, each MEC server is managed by a dedicated DQN agent and DDPG agent, where each DDPG agent comprises an actor, a critic, a target actor and a target critic components.
A transformer-based model is leveraged to better adapt to the variations in task arrivals as a supplement to assist the multi-agent DRL-based algorithm in learning optimal and long-term strategies.

\vspace{-0.1cm}
\subsection{Transformer Prediction Model}
\vspace{-0.1cm}

    We propose using a transformer-based model to predict the future dynamic task arrival time and computational resource requirement of all MEC servers, assisting the MADQN-MADDPG in learning better strategies and optimising long-term goals.
The transformer prediction model is trained offline using historical records of task arrival time and computational requirements. Subsequently, the model migrates time series predictions $Q_{Mt}$ of both the next task arrival time and the upcoming task requirements to the global observation state $s_t$ in MADQN-MADDPG. The predicted task arrival time and resource requirements $Q_{Mt}$ generated by the transformer model are integrated into the global state used by both the MADQN agent and the DDPG critic at each time slot $t$. This integration allows the learning algorithms to better anticipate future demands and optimise decision-making by effectively capturing the temporal dependencies inherent in dynamic task arrivals. As a result, the  overall learning process is enhanced and the corresponding efficiency of resource allocation is improved in the MEC system.

\subsection{Multi-Agent Markov Decision Process}
Considering the long-term resource dynamics in the network, to optimise task offloading and resource management, we formulate Problem~(\ref{objective}) as an MDP defined by a tuple $V = \langle S, A, F, R, \gamma\rangle$, in which $S, A, R, \gamma$ represent state space, action space, reward, and discount factor (to balance the influence of future rewards), respectively. The transition function ${F}$ defines the probability of moving from a given state $s_t\in {S}$ to any state $s_{t+1} \in S$, contingent upon taking an action $a_{t} \in {A}$.
The key elements of the MDP are detailed as follows:

\textit{State of DDPG actor and critic $S$:} Each critic has a global observation view as well as the transformer predictions information, accessing the resource information and the requests of the entire edge network. The state of a critic $s_{t}$ can be written as: 
\begin{equation}
\vspace{-0.1cm}
s_t= [C_{Mt},  B_{Kt} , D_{Mt}, Q_{Mt}],
\end{equation}

where $C_{Mt}$ and $B_{Kt}$ denote the remaining resource of all MECs and links over the network at time $t$, respectively. Furthermore, $D_{Mt}$ and $Q_{Mt}$ represent computing resource requirements and transformer predictions of future task arrivals time and requirements at time slot $t$, respectively.
In contrast, the actor possesses only a partial observation of the state $s_{mt}$, consisting of the remaining resource information of the local MEC server, links to adjacent MEC servers, and details regarding the task size and resource requirements of the local MEC server. Notably, the transformer predictions of future task arrival time and computational resource demands are not integrated into the actor state. The state of actor $s_{mt}$ is:
\begin{equation}
s_{mt}= [C_{mt},  B_{kt} , D_{mt}]  \ \ \forall m \in M_{t},  \forall k \in K_{t}
\end{equation}

\textit{State of DQN:} Each DQN agent shares the same state of DDPG actor $s_{mt}$, which only has a partial view.

\textit{Action of DQN agent:} Action $j_{mt}$ can be represented as: 
\vspace{-0.1cm}
\begin{equation}
j_{mt} = [u_{mt},w_{mt}]  \ \ \forall m \in M_t,
\end{equation}
\vspace{-0.1cm}
where $u_{mt}$ is the selected number of offloading target cooperative MEC servers and $w_{mt}$ represents the selected target MEC servers.

\textit{Action of DDPG actor:} Action $a_{mt}$ can be obtained as:
\vspace{-0.2cm}
\begin{equation}
a_{mt} = [a_{mdt} * D,  \kappa a_{mct} * C,  \kappa a_{mkt}* B]  \ \ \forall m \in M_{t}, \forall k \in K_{t}
\end{equation}

We assume that task size and computational resource requirements are always offloaded with equal proportions. \( a_{mdt} \) represents the proportion of task offloading, which directly affects the required computing resources. \( a_{mct} \) and \( a_{mkt} \) denote the allocated computing and bandwidth resources, respectively. These parameters can be further formulated as follows:
\begin{equation}
    a_{mct} = \max\left( 0, \min\left( a_{mct} + \epsilon \mathcal{O}(\mu, \sigma, \beta), 1 \right) \right)
\end{equation}
\begin{equation}
    a_{mkt} = \max\left( 0, \min\left( a_{mkt} + \epsilon \mathcal{O}(\mu, \sigma, \beta), 1 \right) \right)
\end{equation}

\noindent where \( \mathcal{O} \) represents the Ornstein–Uhlenbeck (OU) noise, and \( \epsilon \) is the noise scale factor. Parameters \( \mu \), \( \sigma \), and \( \beta \) denote the long-term mean, standard deviation, and rate of mean reversion of the OU process, respectively. The limit factor \( \kappa \) is used to limit the maximum resources that can be allocated to each MEC.

\textit{Reward $R$:} Consistent with the objective function of formula~(\ref{objective}), the system reward at time slot $t$ is achieved by: 
\begin{equation}
r_{t} =  \lambda L^* / L_{\text{total}} + \rho E^* / E_{\text{total}}
\end{equation}

\noindent where the parameters $L^*$ and $E^*$ are derived by calculating the average of multiple minimum values within a sliding window that records the values of $L_{\text{total}}$ and $E_{\text{total}}$ over a recent time period. It is implemented to mitigate the impact of outliers, thereby enhancing the algorithm's overall stability.

\begin{figure}[t]  
    \centering
    \includegraphics[width=0.49 \textwidth]{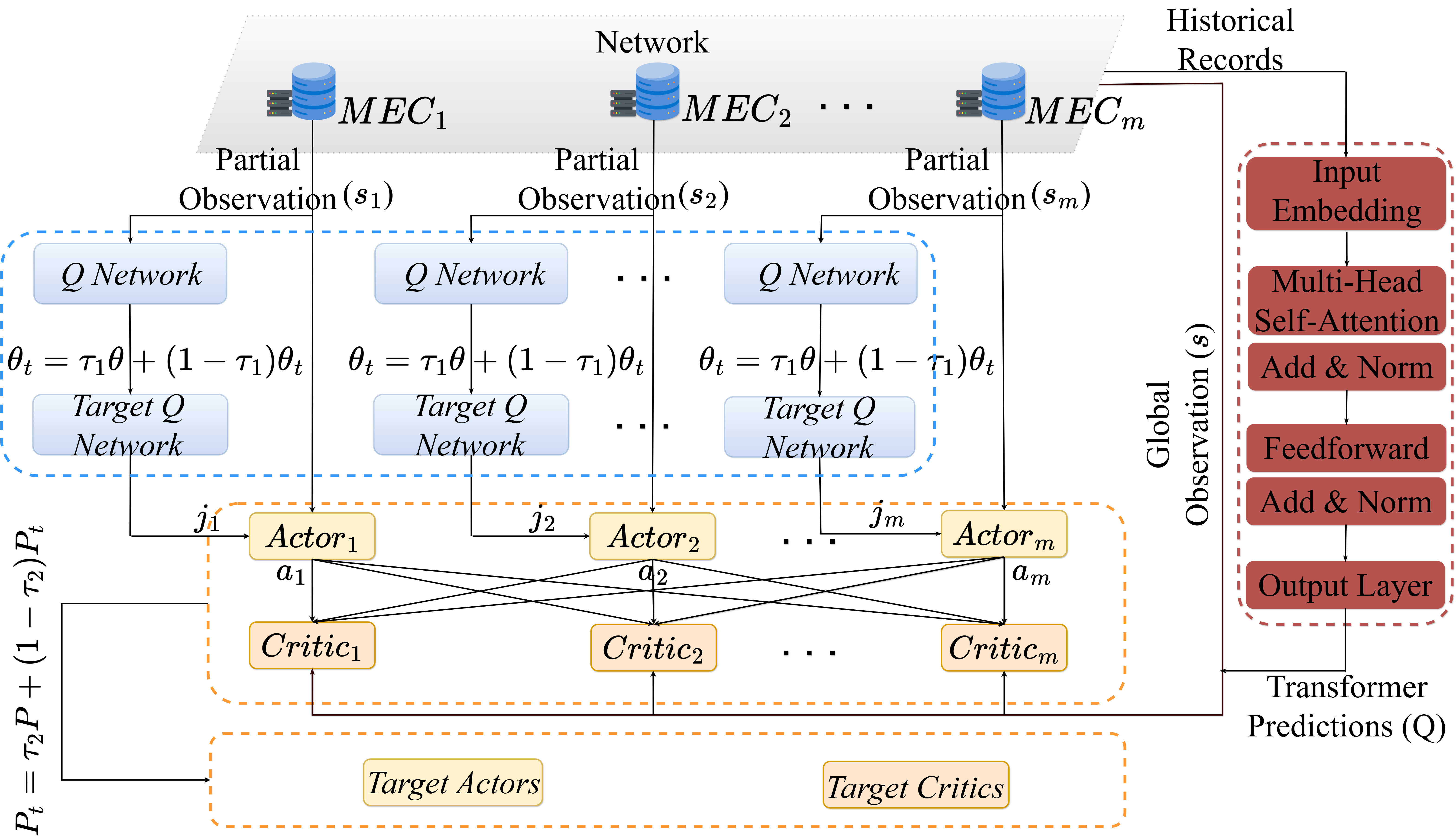} 
    \caption{Proposed Transformer-Driven Hybrid MADQN-MADDPG approach.}
    \label{DRLmodel}
\end{figure}

\subsection{Asynchronous Hybrid MADQN-MADDPG method}
As the action space $a_t = [j_{mt}, a_{mt}]\in {A},\ \forall m \in M_{t}$ comprises both discrete domain and continuous domain, it is divided into two parts, the discrete part $j_{mt}$ represents server selection and the continuous part $a_{mt}$ contains the proportion of offloaded tasks, the allocated computational and bandwidth resources of the selected MEC servers.
The implementation details of the transformer-driven Hybrid MADQN-MADDPG algorithm are illustrated in Fig. \ref{DRLmodel}. The DQN agent captures a partial observation state from each MEC server and then creates an action $j_{mt}$ based on the state. In addition, the action $j_{mt}$ will be diverted to the DDPG's actor, which together with the same partial observation from each MEC server will be used as a reference to choose the actor's action. Specifically, in the common DDPG framework, the critic receives a global observation state representation, and we extend this by incorporating future resource requirements predictions generated by the Transformer model as additional inputs to better capture the temporal characteristics of the dynamic environment. Furthermore, in DRL, both DQN and DDPG use the soft update to enhance model stability and prevent significant oscillations between the policy and target networks. As for DQN, the purpose of the soft update is to allow the target Q-network $\theta_t$ to gradually converge towards the parameters of the current Q-network $\theta$ rather than copying the Q-network parameters directly to the target network at each time step. The soft update process can be written as:
\vspace{-0.1cm}
    \begin{equation}
    \theta_t = \tau_1 \theta + (1 - \tau_1) \theta_t
    \vspace{-0.1cm}
    \end{equation}

\noindent where $\tau_1$ is the soft update coefficient, typically between 0 and 1.
Similarly in DDPG, soft updates are applied to the target network of actor and critic, which is represented as:
\vspace{-0.1cm}
    \begin{equation}
    P_t = \tau_2 P + (1 - \tau_2) P_t
    \vspace{-0.1cm}
    \end{equation}

\noindent where $P_t$ and $P$ are the parameters of the target network and main network. $\tau_2$ represents the soft update coefficient of DDPG actor and critic network.

In each training step of the proposed asynchronous hybrid MADQN-MADDPG, only agents with task requests make actions based on partial observation and train based on global observation. The action is set to $0$ when the agent has no task requests to process, which helps other agents implement the strategies effectively. Each DDPG agent stores experience in both the independent replay buffer and the global replay buffer. During asynchronous training, agents sample experience from their independent replay buffers but periodically synchronise by extracting experiences from the global replay buffer containing the experience of all agents. Conversely, DQN agents operate without needing periodic synchronisation because they collectively utilise and train a shared Q-Network, thereby inherently integrating their experiences throughout the asynchronous training.
Accordingly, the asynchronous hybrid MADQN-MADDPG algorithm achieves the combination of discrete and continuous action space, optimising the server selection, task offloading proportion, and resource allocation ratio. This framework reduces the synchronous waiting time by training asynchronously while ensuring active collaboration across MEC servers, thereby comprehensively adapting to dynamic distributed edge network environments.


\section{Performance Evaluation}
\label{sec:results}


\begin{table}[b]
  \vspace{-0.5cm}
  \raggedright
  \caption{Simulation Parameters}
  \vspace{-0.2cm}
  \renewcommand{\arraystretch}{0.9} 
  \resizebox{\columnwidth}{!}{ 
    \begin{tabular}{|l|l||l|l|}
    \hline
    \textbf{Parameters} & \textbf{Value}  & \textbf{Parameters} & \textbf{Value} \\ \hline
    CPU clock speed (computing capacity) & 100 GHz &  Batch size  & 300  \\ \hline
    Bandwidth capacity $B$ & 10 Gbps & Update coefficient $\tau$ & 0.005  \\ \hline
    Noise-to-signal ratio $\eta$ & 10 dB & DRL Discount factor  & 0.99  \\ \hline
    Initial scale $\epsilon$ & 1 &  Limit factor $\kappa$    & 0.4  \\ \hline
    Latency Weight $\lambda$ &  0.5             & Long-term mean $\mu$    & 0  \\ \hline
    Energy Weight $\rho$     &  0.5      & Reversion rate $\beta$ &   1 \\ \hline
    Learning rate   & 0.0005      & Standard deviation $\sigma$&  0.3 \\ \hline
    \end{tabular}
  }
  \label{table:parameters}
\end{table}

In this section, we evaluate the performance of the proposed 
CTO-TP algorithm based on the asynchronous hybrid MADQN-MADDPG. We consider 
a MEC network comprising three servers, where all the MEC servers have task requests and can offload all or partial tasks to each other. The length of one training step is 1 second and the task requests are processed at each training step. If there is no task request for all MEC servers in one training step, this training step will be skipped. According to \cite{li2023drl}, MEC servers' minimum power $P_{min}$ and maximum power $P_{max}$ are typically 176 W and 396 W, respectively. This paper uses the Google Cluster Traces dataset \cite{reiss2011google}, which includes task arrival time and computational resource requirement. The size of each task is randomly produced between 5 and 20 MB because this dataset excludes it. Transformer-based model is configured with model dimension, attention heads, encoder layer, and feedforward dimension, which are set to 512, 4, 1, and 1024, respectively. Other simulation parameter values are detailed in Table \ref{table:parameters}. The selection of these specific coefficients was determined empirically during the experimental research to generate optimal performance values. The comparison solutions used in the task offloading simulation are:
\vspace{-0.1cm}
\begin{itemize}
    \item \ \textit{Cooperative task offloading without transformer-driven prediction (CTO)}: In this approach, the transformer-based prediction model is removed from CTO-TP algorithm to evaluate the impact of incorporating task arrival time and computational resource requirement predictions provided by the transformer-based model.
    \item \ \textit{Full Allocation(FA)}: FA is a heuristic resource allocation method that seeks to maximise the utilisation of available computational and bandwidth resources by fully allocating resources without considering specific system states and optimisation goals.

    \item \ \textit{Random Allocation(RA)}: RA is a task offloading and resource management solution where all the allocation decisions are made randomly within defined constraints.

\end{itemize}

\begin{figure}[t]
    \centering
    \begin{subfigure}{0.24\textwidth}
        \centering
        \includegraphics[width=\textwidth]{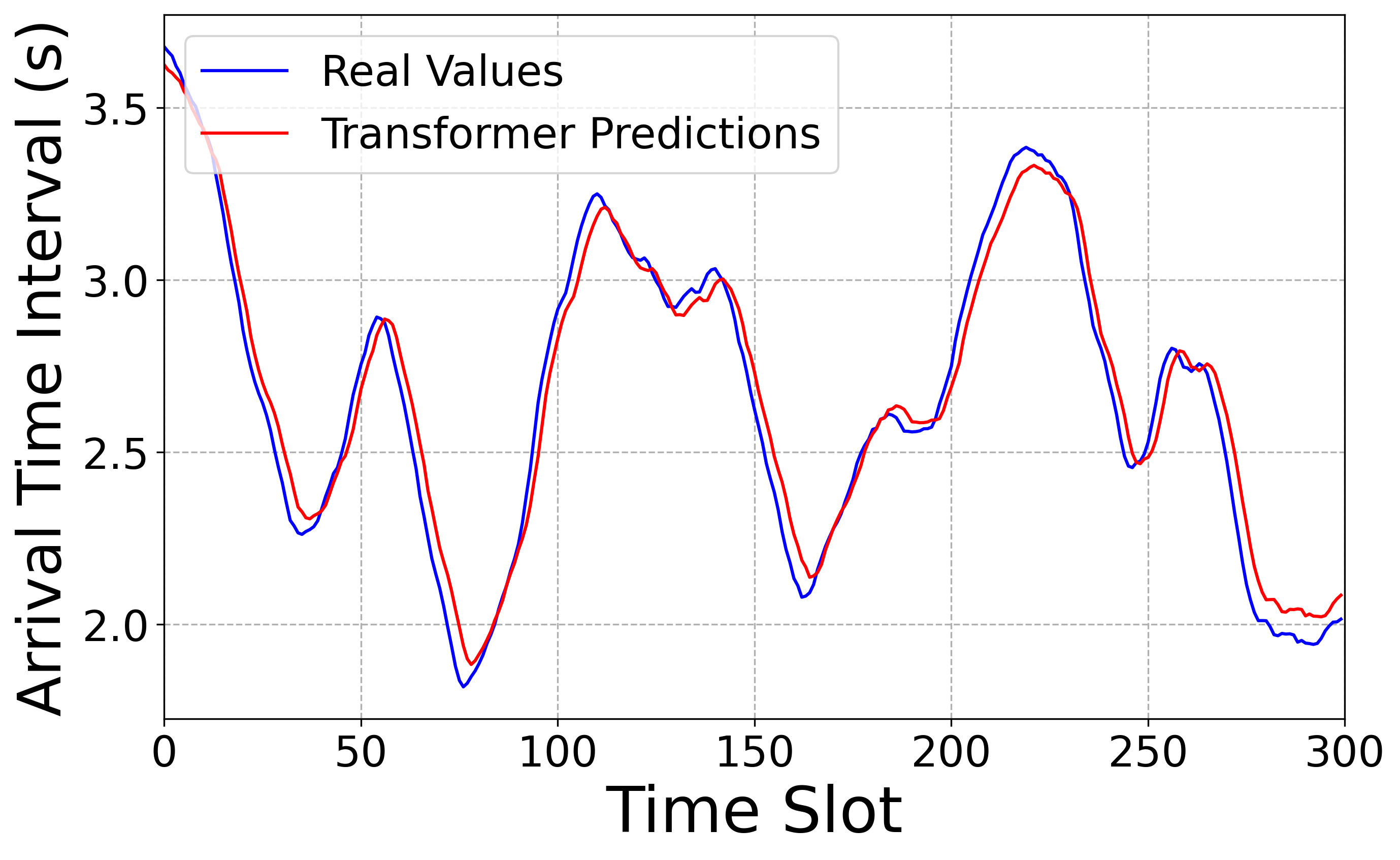}
        \vspace{-0.6cm}
        \caption{}
        \label{fig:timeinterval}
    \end{subfigure}
    \hfill
    \begin{subfigure}{0.24\textwidth}
        \centering
        \includegraphics[width=\textwidth]{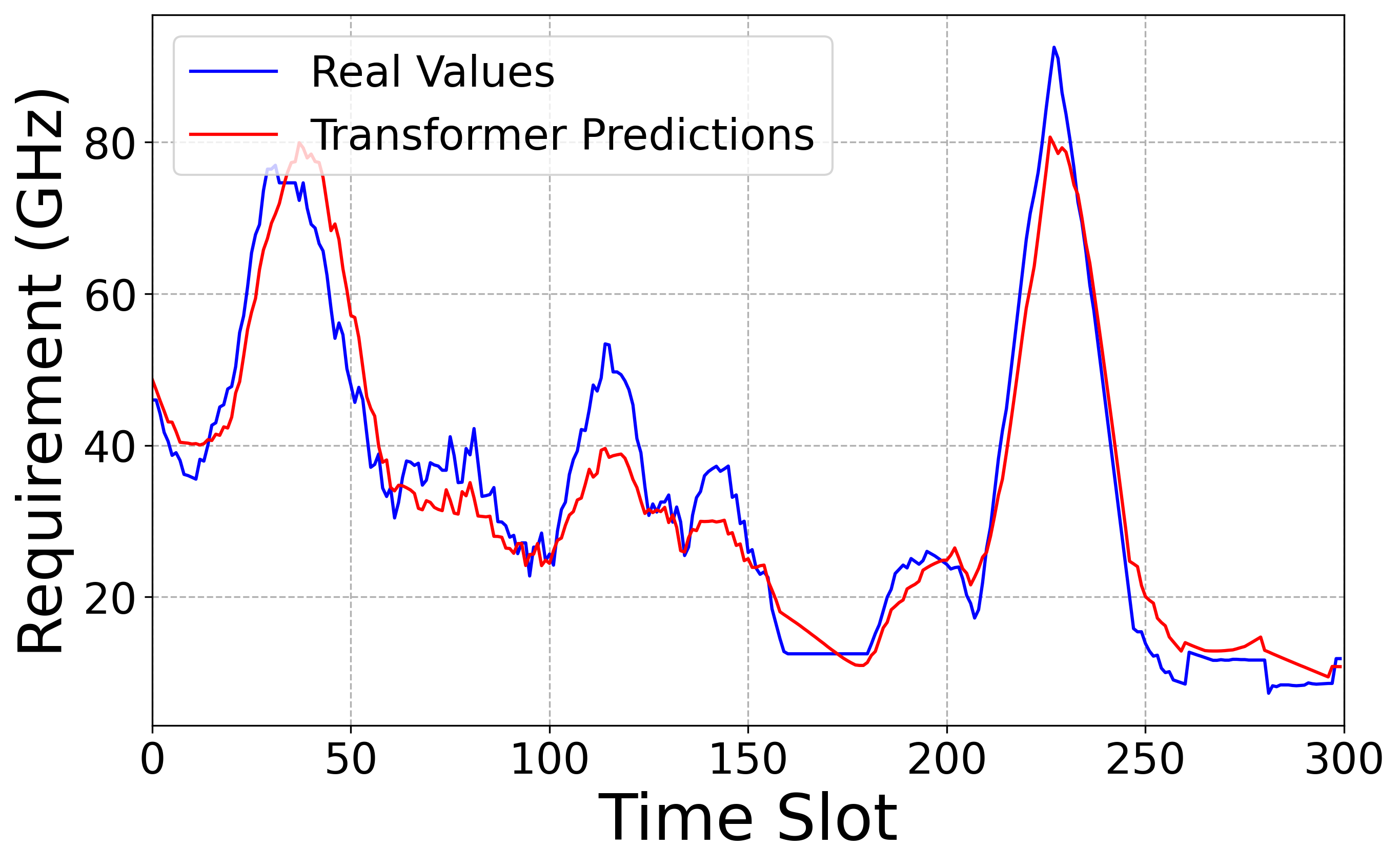}
        \vspace{-0.6cm}
        \caption{}
        \label{fig:requirements}
    \end{subfigure}
    \vspace{-0.7cm}
    \caption{Comparison of real values and predicted values by transformer-based models, (a) Task Arrival Time Interval, (b) Task Computing Resource Requirement.}
    \vspace{-0.5cm}
    \label{fig:prediction}
\end{figure}

The coefficient of determination $R^2$ is used in the experiment to evaluate the transformer time-series prediction model's performance, and all $R^2$ scores are capable of reaching above $0.90$. The task arrival time interval and task requirement represent the time interval between the current task and the next task, and the computing resource requirement of the current task.
As shown in Fig. \ref{fig:prediction}, the transformer-based model can predict trends and values of task arrival time intervals and requirements effectively, especially in maintaining high performance during the peak task requirements. Figure \ref{fig:prediction} is achieved by employing a sliding window of size 20 to smooth fluctuations.

\begin{figure}[t]  
    \centering
    \includegraphics[width=0.4\textwidth]{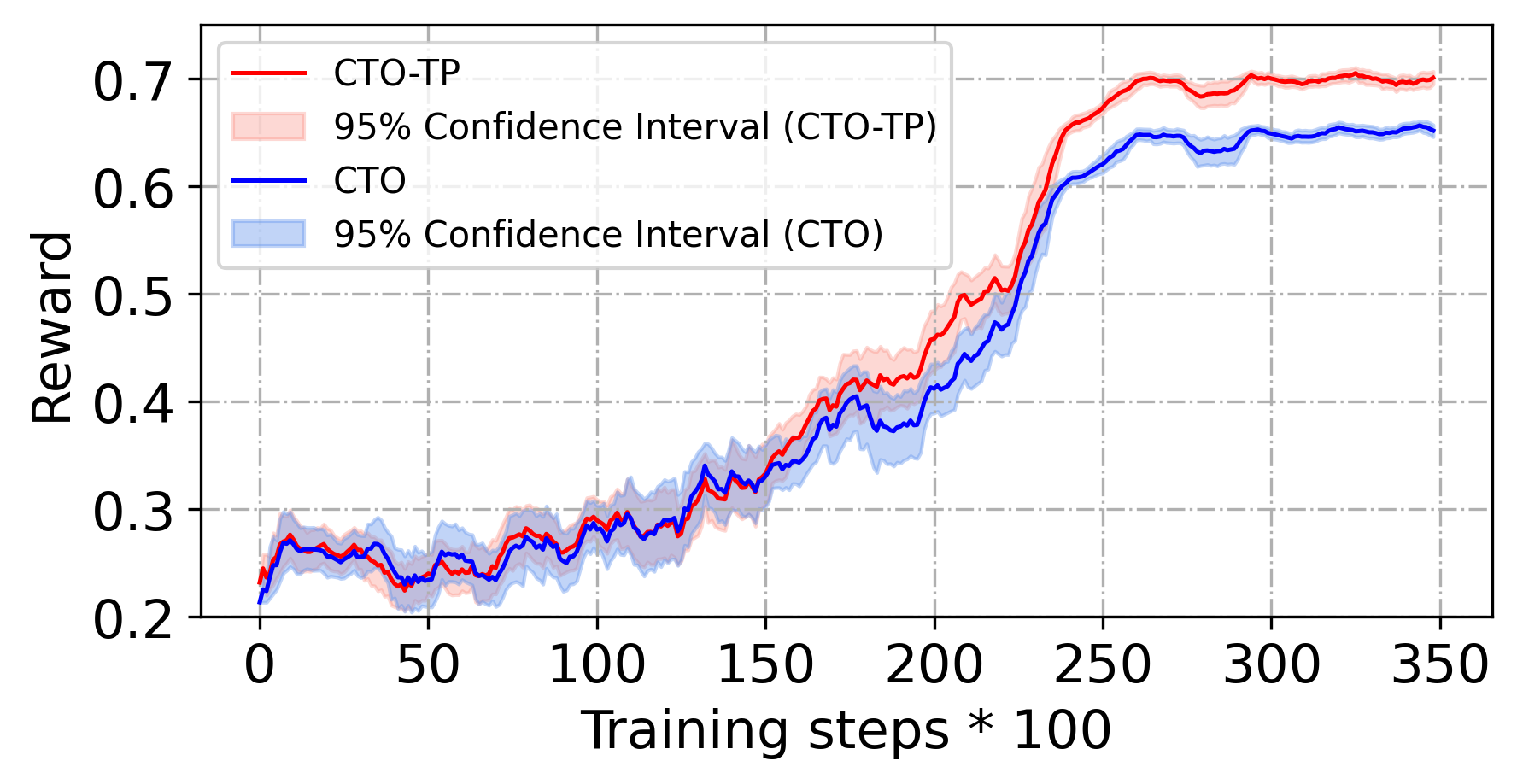}  
    \vspace{-0.3cm}
    \caption{Performance evaluation of CTO-TP and CTO.}
    \label{fig:CTO-TP&CTO}
    \vspace{-0.7cm}
\end{figure}

\begin{figure}[t]  
    \centering
    \includegraphics[width=0.4\textwidth]{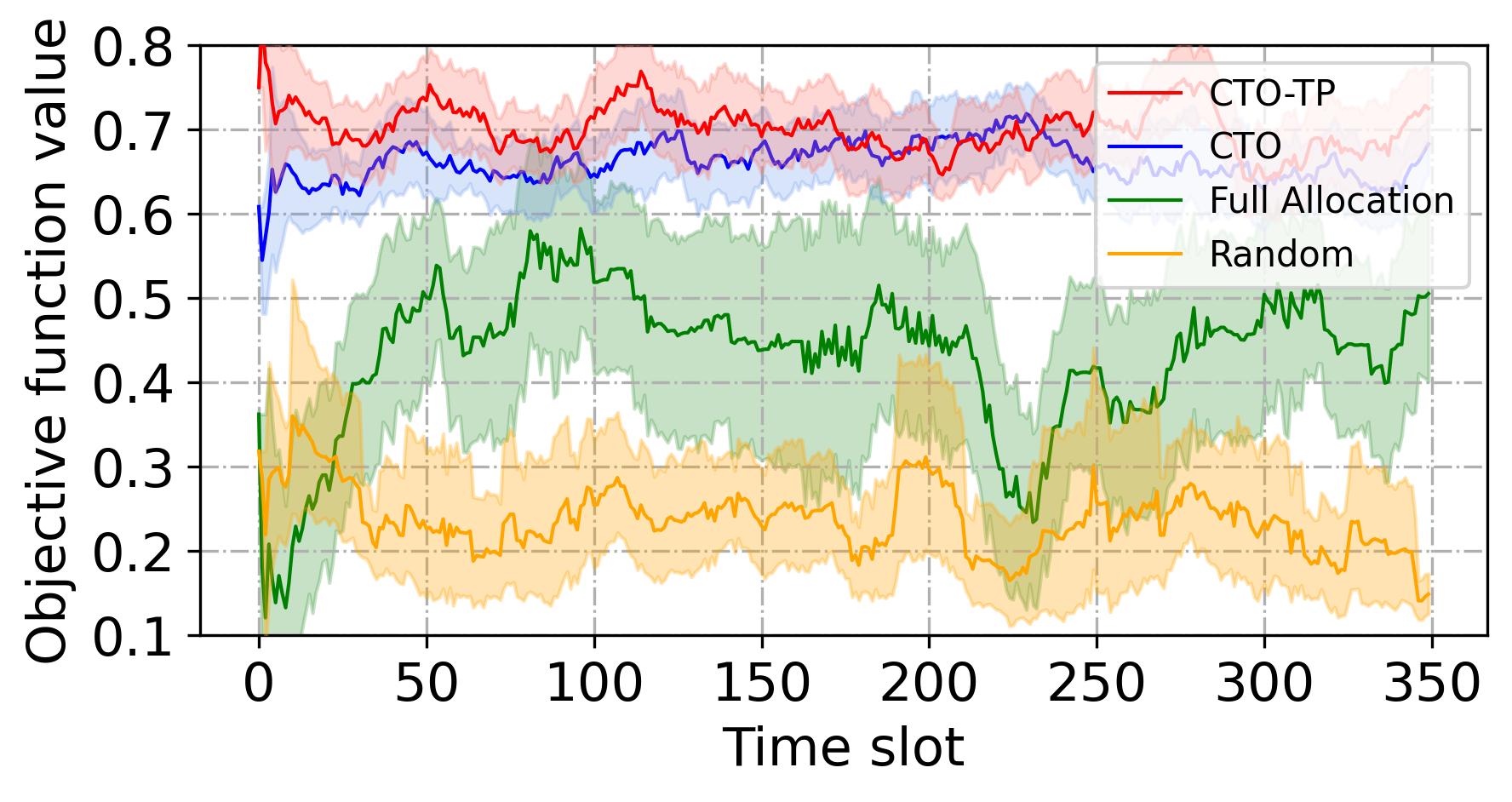} 
    \vspace{-0.3cm}
    \caption{Performance comparison of various task offloading and resource management solutions.}
    \label{fig:benchmark}
    \vspace{-0.6cm}
\end{figure}

\captionsetup[subfigure]{labelformat=simple, labelsep=period}
\renewcommand\thesubfigure{(\alph{subfigure})}

\begin{figure}[t]  
    \centering
    \includegraphics[width=0.4\textwidth]{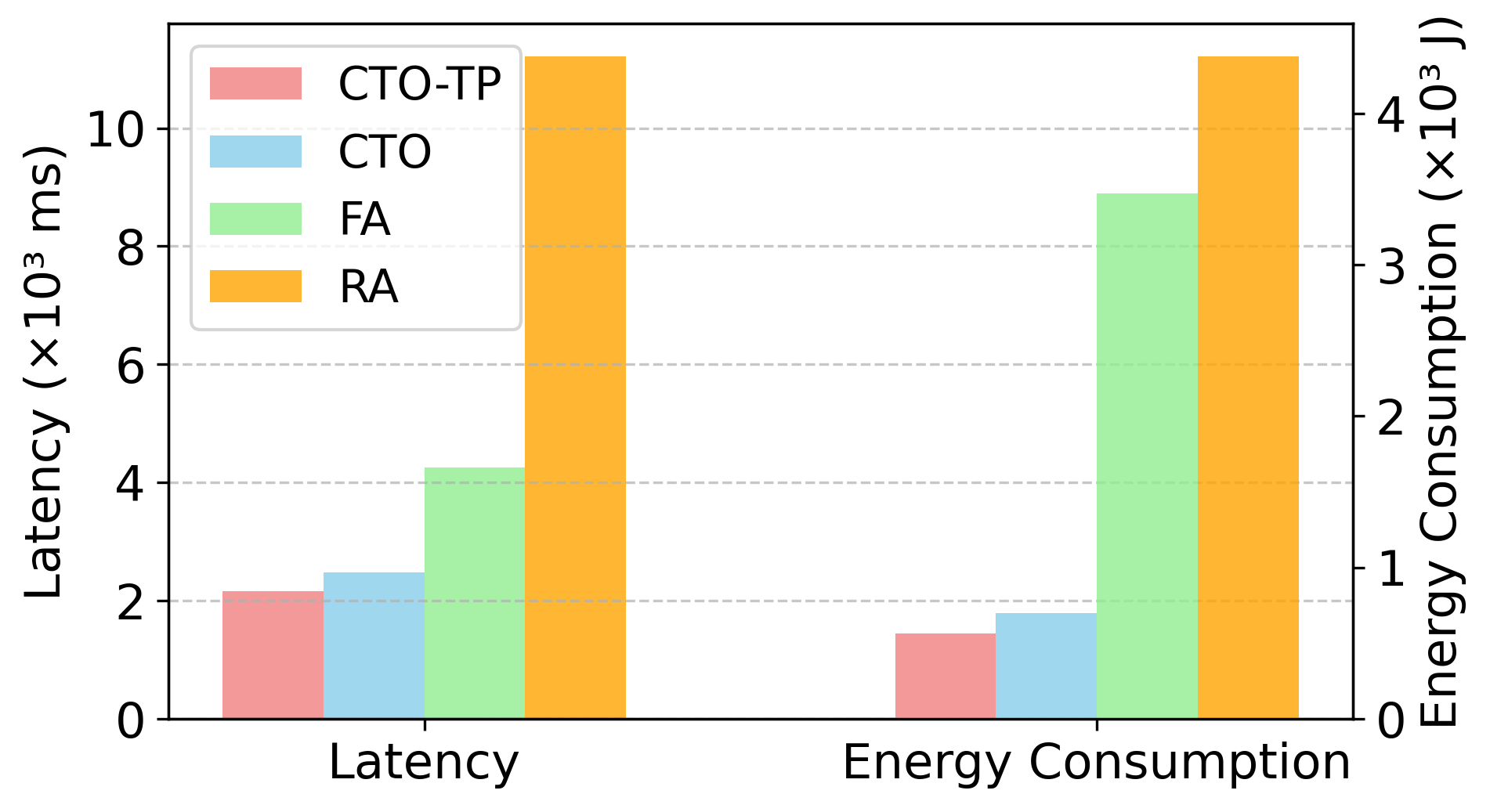} 
    \vspace{-0.2cm}
    \caption{Performance of different methods for Average System Latency and Average System Energy Consumption.}
    \label{fig:LatencyEnergy}
    \vspace{-0.7cm}
\end{figure}

 Figure  \ref{fig:CTO-TP&CTO} shows a comparative analysis of the training process between CTO-TP and CTO algorithms. The observation is conducted before 5000 training steps, while CTO-TP and CTO start training after this period, proceeding till the end. Both methods ultimately converge, achieving reward values of 0.70 and 0.66, respectively. The prediction of task arrival time and requirements from the transformer-based model facilitates hybrid MADQN-MADDPG agents to perform more accurate task offloading and resource allocation, thereby enhancing their policy optimisation. Furthermore, the actual execution time of these optimisation methods is crucial to task offloading. We conducted a large number of experiments in both CTO-TP and CTO, and obtained the actual execution time of both CTO-TP and CTO is 1.6 s. Nevertheless, CTO-TP has higher performance in terms of latency and energy consumption.
Similarly, Fig. \ref{fig:benchmark} illustrates the performance comparison among various task offloading and resource management solutions. The RA solution yields the lowest performance, indicating that random allocation leads to inefficient resource utilisation. The FA method, while an improvement over RA, achieves only moderate enhancement by ensuring full resource allocation and utilisation, with the lack of adaptation to system states, dynamic task requests, and optimisation goals. Conversely, the CTO and CTO-TP algorithms exhibit significantly higher performance, as actions are consistently aligned in consideration of dynamic task requests and real-time system conditions. This dynamic adaptability promotes sufficient collaboration among all agents, maximising long-term gains across multiple optimisation objectives. In addition, the CTO-TP method surpasses CTO in performance due to the benefits of accurate predictions of task arrival time and task requirements in the future.

 Figure \ref{fig:LatencyEnergy} presents a comparison of average system latency and energy consumption (i.e., the primary optimisation objectives cross the CTO-TP), CTO, FA, and RA benchmarks. The random resource allocation incurs the highest average latency and energy consumption attributed to the imbalanced allocation of computing and bandwidth resources under the random strategy. Although the FA management enhances system resource efficiency over RA by ensuring a full resource utilisation policy, it remains ineffective in adapting to the dynamic environment. In contrast, the CTO-TP and CTO accomplish lower latency and energy consumption levels, benefiting from adaptive adjustments in task offloading and resource allocation strategies for computing and bandwidth resources. These adjustments respond to the system's dynamic conditions, comprising varying task arrival times and requirements, coupled with remaining computational and communicational resource fluctuation. Furthermore, CTO-TP algorithm leverages its predictive capability for future task arrival time and computational resource demands to enable the system to adjust resource management strategies in advance, thereby achieving the lowest system average latency and energy consumption, ultimately enhancing the quality of service.

\section{Conclusion and Future Work}
\vspace{-0.1cm}
\label{sec:conclusion}
In this paper, we investigated the problem of cooperative task offloading in Mobile Edge Computing. To address this problem, a novel joint latency and energy-efficient cooperative task offloading mechanism with transformer-driven prediction algorithm, based on the asynchronous hybrid MADQN-MADDPG, is proposed. This method optimises the task offloading strategies and resource allocation management while reducing the synchronous waiting time by performing asynchronous training on the foundation of enhanced edge-edge cooperation between edge servers. This approach minimises the long-term system latency and energy consumption. Verification of the proposed scheme through testbed experiments and further enhancement is planned as future work. 

\vspace{-0.1cm}

\printbibliography

\end{document}